\documentclass[11pt]{article}

\usepackage{emnlp2021}
\usepackage{setspace}
\usepackage{times}
\usepackage{graphicx}
\usepackage{xspace}
\usepackage{subfigure}
\usepackage{latexsym}
\usepackage{enumitem}
\usepackage{booktabs}
\usepackage{tabularx}
\usepackage{amsmath}

\usepackage{microtype}

\newcommand{\our}{\mbox{\texttt{AutoKeyGen}}\xspace}

\newcommand{\smallsection}[1]{\noindent\textbf{#1.}}

\title{Unsupervised Deep Keyphrase Generation}
\author{

Xianjie Shen $^1\;\;\;\;\;$ Yinghan Wang $^3\;\;\;\;\;$  Rui Meng$^4\;\;\;\;\;$ Jingbo Shang $^{1,2}$ \\
\small $^1$ Department of Computer Science and Engineering, University of California San Diego, CA, USA \\
\small $^2$ Hal\i c\i o\u glu Data Science Institute, University of California San Diego, CA, USA \\
\small $^3$ Department of Computer Science, University of Virginia, VA, USA \\
\small $^4$ Department of Computer Science, University of Pittsburgh, PA, USA \\
\small \texttt{\{xishen, jshang\}@ucsd.edu $\qquad$ yw9fm@virginia.edu $\qquad$ rui.meng@pitt.edu }\\
}

\date{}

\begin{document}

\maketitle

\begin{abstract}
    Keyphrase generation aims to summarize long documents with a collection of salient phrases. 
% Previous unsupervised methods have achieved decent performance but can hardly predict absent keyphrases.
Deep neural models have demonstrated a remarkable success in this task, capable of predicting keyphrases that are even absent from a document. However, such abstractiveness is acquired at the expense of a substantial amount of annotated data. 
In this paper, we present a novel method for keyphrase generation, \our, without the supervision of any human annotation.
Motivated by the observation that an absent keyphrase in one document can appear in other places, in whole or in part, we first construct a phrase bank by pooling all phrases in a corpus.
% We observe that absent keyphrases in one document could be present in other documents.
% Absent keyphrases could also partially appear in the input document, maybe as separate tokens.
% We design a tailored noun phrase extractor to all documents for present candidates and pool them together into a phrase bank.
% Therefore, we follow the literature to extract candidate present keyphrases from all documents and then pool them together into a phrase bank. 
%\jingbo{I think we should tone down (or even remove it here in abstract) this noun phrase extractor and put more emphasis on the phrase bank. }
With this phrase bank, we then draw candidate absent keyphrases for each document through a partial matching process.
To rank both types of candidates, we combine their lexical- and semantic-level similarities to the input document.
Moreover, we utilize these top-ranked candidates as  to train a deep generative model for more absent keyphrases. 
Extensive experiments demonstrate that \our outperforms all unsupervised baselines and can even beat strong supervised method in certain cases.
 
\end{abstract}
\begin{figure*}[t]
 \centering
    \includegraphics[width=\linewidth]{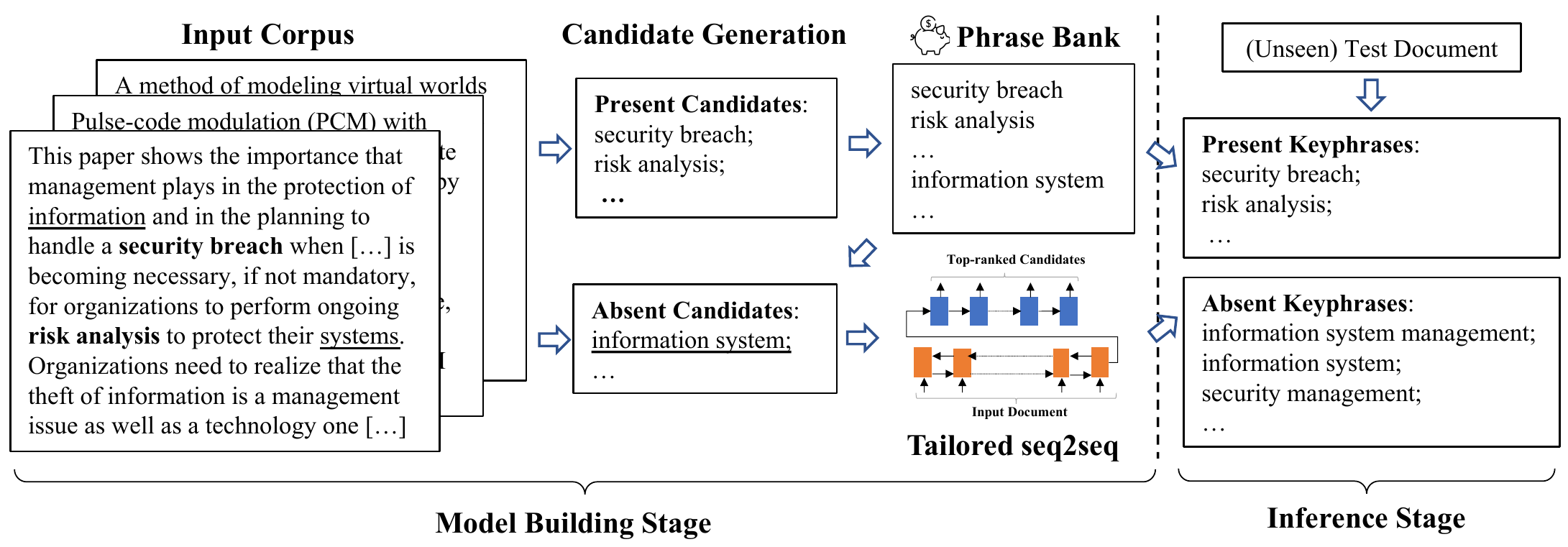}
    \vspace{-3mm}
    \caption{An overview of our proposed \our framework with a part of real example. The full version of the example can be found in our case study.}
    \label{fig:overview}
    \vspace{-3mm}
\end{figure*}

\section{Introduction}
\label{sec:intro}

Keyphrase generation aims to produce a list of short phrases to summarize and characterize a long document (e.g., research papers and news articles).
It has a wide spectrum of applications, to name a few, information retrieval~\citep{Jones1999PhrasierAS}, text summarization~\citep{10.5555/1039791.1039794}, and text categorization~\citep{hulth-megyesi-2006-study}.

The trade-off between the capability of generating absent keyphrases (i.e., phrases do not appear in the original document) and the reliance on document-keyphrase supervision has long existed among keyphrase generation methods.

Extractive methods~\cite{hasan2010conundrums,Shang_2018,DBLP:conf/conll/Bennani-SmiresM18} can only predict phrases that appear in the original document. Nevertheless, many of them do not need any direct supervision and they demonstrate great robustness across various genres of text.
Some studies expand the extraction scope from the input document to its neighbor documents~\cite{DBLP:conf/aaai/WanX08, florescu-caragea-2017-positionrank}, but they still cannot predict absent keyphrases well. 
~\citet{Meng_2017} has shown that in scientific documents, up to 50\% of keyphrases are absent from the source text, yet they can be helpful for applications such as search and recommendation~\cite{boudin2021redefining}.

With the advance of deep neural networks, recent studies~\cite{Meng_2017,chen2019guided,sun2019divgraphpointer,alzaidy2019bi,yuan2018size,meng2020empirical} are capable of generating keyphrases, according to their semantic relevance to a document, no matter they are present or not.
Although these methods have achieved state-of-the-art performance,
all these deep models are supervised and typically require a tremendous number of document-keyphrase pairs, which could be expensive and laborious to collect. 
For example, \citet{Meng_2017} utilized more than 500,000 author-annotated scientific papers to train a RNN model.
Similarly, \citet{xiong2019open} collected 68,000 webpages and have them annotated by professional annotators. 

In this paper, we aim to alleviate this trade-off by proposing an unsupervised method that can generate both present and absent keyphrases without utilizing any human annotations. We observe that absent keyphrases of a document can be present in other documents as present keyphrases.
Also, many absent keyphrases in fact appear in the original document in part as separate tokens. 
% We use the Inspec dataset, one of the benchmark datasets in keyphrase generation, as an example.
For example, in the Inspec dataset, one of the benchmark datasets in keyphrase generation,
% In this dataset, 
99\% of absent keyphrases can be found in other documents. And for 56.8\% of absent keyphrases, all their tokens separately appear in the input document.

Inspired by these observations, 
we propose a novel unsupervised deep keyphrase generation method \our as illustrated in Figure~\ref{fig:overview}.
Specifically, we first follow previous works~\cite{hasan2010conundrums,Shang_2018,DBLP:conf/conll/Bennani-SmiresM18} to extract candidate present keyphrases from all documents and then pool them together into a phrase bank.
% we first extract present candidates from each document similar to existing work~\cite{hasan2010conundrums,Shang_2018,DBLP:conf/conll/Bennani-SmiresM18} and then pool them together into a phrase bank.
% We observe that the absent keyphrases appear as present keyphrases in other documents quite often and the words in those absent phrases may exist in the input document although not continuously.
From this present phrase bank, we can now draw candidate absent keyphrase for each document through a partial matching process, requiring each stemmed word in the candidate phrase should exist in the input document. 
To rank both types of keyphrases, we fuse two popular measurements in unsupervised keyphrase extraction methods, i.e., the TF-IDF score at the lexical level and embedding similarity at the semantic level.
We further utilize these top-ranked present and absent candidates as ``silver'' data to train a deep generative model.
This generative model is expected to augment absent keyphrases by a biased beam search method, which encourages the model to predict words from the input document instead of from the vocabulary.

Extensive experiments show that \our outperforms all unsupervised baselines consistently, and even the strong supervised baseline in certain cases.

Our contributions are summarized as follows:
\begin{itemize}[nosep,leftmargin=*]
    \item We make two important observations about absent keyphrases, illuminating the feasibility of training abstractive keyphrase models in an unsupervised manner.
    % analyze the trade-off between the capability of generating absent keyphrases and the reliance on annotated supervision, and 
    \item We propose a novel unsupervised deep keyphrase generation method \our that can perform well on predicting both present and absent keyphrases.
    \item We conduct extensive experiments on five benchmark datasets and demonstrate the superiority of our method \our over unsupervised baselines. On some datasets, \our even yields better results than state-of-the-art supervised methods.
\end{itemize}

\smallsection{Reproducibility} We release our codes and datasets on GitHub.\footnote{\url{https://github.com/Jayshen0/Unsupervised-Deep-Keyphrase-Generation}}
%\jingbo{replace this URL by the real one}

\section{Problem Formulation}

In this work, we aim to build a keyphrase generation model solely based on a collection of documents $\mathcal{D}$, without any keyphrase annotation.
Keyphrase generation is typically formulated and evaluated as a ranking problem.
Given an (unseen) input document $\mathbf{x}$, the goal of this task is to output a ranked list of keyprhases $\mathcal{Y}$.
We denote the input document as a sequence of tokens, i.e., $\mathbf{x}=[x_1, \ldots, x_{|\mathbf{x}|}]$. 
Here, $|\mathbf{x}|$ is the total number of tokens in this document. 

Depending on whether a keyphrase appears in the input document or not \emph{as a whole unit}, one can categorize the keyphrases in $\mathcal{Y}$ into two ranked lists: 
(1) Present keyphrase ranked list, $\mathcal{Y}^P=\{\mathbf{y}^p_1, \ldots, \mathbf{y}^p_{|\mathcal{Y}^P|}\}$
and (2) Absent keyphrase ranked list: $\mathcal{Y}^A=\{\mathbf{y}^a_1, \ldots, \mathbf{y}^a_{|\mathcal{Y}^A|}\}$.
Here, $|\mathcal{Y}^P|$/$|\mathcal{Y}^A|$ is the number of present/absent keyphrase predictions respectively.
That is, $\mathcal{Y} = <\mathcal{Y}^P, \mathcal{Y}^A>$.
Each keyphrase is also a sequence of tokens, which can contain single or multiple tokens.
\section{Our \our Method}

\smallsection{Overview}
As shown in Figure~\ref{fig:overview}, the training process of \our consists of three steps:
(1) pool the candidate present keyphrases from all documents together as a phrase bank and then draw candidate absent keyphrases for each document;
(2) rank all these candidates based on TF-IDF information and embedding similarity between document and candidate phrase;
(3) train a Seq2Seq generative model using the silver labels derived from the second step to generate more candidate phrases that might be absent in the document or missed in the previous steps. 

When it comes to the inference for new documents, \our will extract candidates following the phrase bank and generate candidates using the Seq2Seq model, and then, rank these candidates together following the same ranking module as (2).

\subsection{Phrase Bank for Absent Keyphrases}
% Similar to existing extractive methods~\cite{hasan2010conundrums,Shang_2018,DBLP:conf/conll/Bennani-SmiresM18}, we use noun phrase extractor to firstly obtain the present candidates for each document. 

\smallsection{Phrase Bank Construction}
As aforementioned, absent keyphrases in one document often appear in other documents.
For example, in the Inspec dataset, one of the benchmark datasets in keyphrase generation,
99\% absent keyphrases are present keyphrases in some other documents; 
Therefore, we first construct a phrase bank by pooling together the present candidates extracted from every document in the raw document collection $\mathcal{D}$.
Specifically, we follow the literature~\cite{hasan2010conundrums,Shang_2018,DBLP:conf/conll/Bennani-SmiresM18} to extract candidate present keyphrases from all documents.
% The details can be found in Appendix.
The details can be found in the implementation details in the experiments section.

\smallsection{Absent Candidate Generation}
In many cases, tokens of an absent keyphrase can in fact be found in the source document but not in a verbatim manner. 
For example, in the Inspec dataset, 56.8\% absent keyphrases have all their tokens separately appeared in the input document.
This inspires us to conduct a partial match as follows.
Given an input document $\mathbf{x}$, one can iterate all phrases in the phrase bank and take as candidates the phrases that all tokens appear in $\mathbf{x}$ (after stemming). 
We enforce the strict requirement of \emph{all tokens} as the phrase bank is huge and there would be too many candidates that can partially appear in $\mathbf{x}$.
For the sake of efficiency, we implement this process via an inverted index mapping document tokens to the phrase bank, so practically we do not have to scan the entire phrase bank for each document.

\subsection{Ranking Module}

The keyphrase generation aims to provide a ranked list of phrases, so we need to rank the obtained candidates.
From the literature, we notice that both lexical and semantic level similarities are important and effective in keyphrase ranking. 
In this paper, we combine both types of similarities.

\smallsection{Embedding Similarity}
According to~\citet{DBLP:conf/conll/Bennani-SmiresM18}, modern embedding methods, such as Doc2Vec~\cite{DBLP:conf/rep4nlp/LauB16}, are able to encode phrases and documents into a shared latent space, then the semantic relatedness can be measured by the cosine similarity in this space.
We follow this work and use the Doc2Vec model pre-trained on the large English Wikipedia corpus to generate 300-dimension vectors for both the input document and its candidate phrases.
Specifically, we denote the embedding of the document $\mathbf{x}$ and the candidate phrase $c$ as $E(\mathbf{x})$ and $E(c)$, respectively.
Their semantic similarity is defined as
\begin{equation*}
\footnotesize
    \mbox{Semantic}(\mathbf{x}, c) = \frac{||E(\mathbf{x})\cdot E(c)||}{||E(\mathbf{x})||\cdot||E(c)||}
\end{equation*}

\smallsection{TF-IDF Information}
TF-IDF, measuring the lexical-level similarity, has been observed as a simple yet strong baseline in literature~\cite{Meng_2017,DBLP:conf/ecir/0001MPJNJ18}.
Specifically, for a document $\mathbf{x}$ in corpus $\mathcal{D}$, the TF-IDF score of phrase $c$ is computed as:
\begin{equation*}
\footnotesize
    \mbox{Lexical}(\mathbf{x}, c)=\frac{\mbox{TF}(c,\mathbf{x})}{|\mathbf{x}|}log\frac{|\mathcal{D}|}{\mbox{DF}(c,\mathcal{D})}
\end{equation*}
where $|\mathbf{x}|$ is the number of word in document $\mathbf{x}$, $\mbox{TF}(c,\mathbf{x})$ is the term frequency of $c$ in $\mathbf{x}$, $\mbox{DF}(c,\mathcal{D})$ is the document frequency of $c$ in $\mathcal{D}$.

\smallsection{Fused Ranking}
We observe that the embedding-based similarity and TF-IDF have different behaviors when the documents are of different lengths. 
Semantic representation learning such as Doc2Vec is reliable for both short and relatively longer documents~\cite{DBLP:conf/rep4nlp/LauB16}.
%\jingbo{I remember some reviewer mentioned some complaint about the previous sentence. Double check that and see if we have fixed it or not.}
TF-IDF works more stable when the document is sufficiently long. 
Therefore, it is intuitive to unify these two heuristics for present keyphrases. 
We propose to combine them using a geometric mean as follows.
\begin{equation*}
\footnotesize
    \mbox{RankScore}(\mathbf{x}, c) =  \sqrt{\mbox{Semantic}(\mathbf{x},c) \cdot \mbox{Lexical}(\mathbf{x},c)}
\end{equation*}
The higher the $\mbox{RankScore}(\mathbf{x},c)$ is, the more likely the candidate phrase $c$ is a keyphrase for the document $\mathbf{x}$.

\subsection{Generation Module}
Using our phrase bank, we can cover more than 90\% of present keyphrases, however, less than 30\% of absent keyphrases are included. 
To bring more absent candidates, we train a Seq2Seq generative model using the highest scored document-keyphrase pairs from the ranking module's results.
Specifically, we pair each document with the top-5 present candidates and top-5 absent candidates, and use these pairs as silver labels for training. 

\smallsection{Classical Encoder-Decoder Model}
The encoder is implemented with BiLSTM~\cite{963769} and the decoder is implemented LSTM. 
%\jingbo{What's the difference between this x and the document x? Are they the same? Also, What's b? Seems some legacy issues for the article body.}
The encoder maps a sequence of tokens in $\mathbf{x}$ to a sequence of continuous hidden representations $(\mathbf{h}_{enc}^1, \ldots, \mathbf{h}_{enc}^{|\mathbf{x}|})$ where $|\mathbf{x}|$ is length of the document, an RNN decoder then generates the target keyphrase~$(y^1, y^2, \ldots, y^{|y|})$ token-by-token in an auto-regressive manner ($|y|$ denotes the number of tokens in the keyphrase):
\begin{equation*}
    \label{eq:decoder}
    \begin{aligned}
        \mathbf{h}^t_{enc} =  f_{enc}(\mathbf{h}_{enc}^{t-1},x^t),\\
        \mathbf{c} = q(h^1_{enc},h^2_{enc},...,h^{|\mathbf{x}|}_{enc}) ,\\
        \mathbf{h}^t_{dec} =  f_{dec}(\mathbf{h}_{dec}^{t-1}, o^{t-1},\mathbf{c})
    \end{aligned}
\end{equation*}
where $\mathbf{h}^t_{enc}$, and $\mathbf{h}^t_{dec}$ are hidden states at time $t$ for encoder and decoder respectively; $f_{enc}$ and $f_{dec}$ are auto-regressive functions implemented by LSTM cells; $o^{t-1}$ is the predicted output of decoder at time $t-1$; and $\mathbf{c}$ is the context vector derived from all the hidden states of encoder though a non-linear function $q$.

At timestep $t$, 
% the next word to predict 
the prediction of $y^t$ is determined based on a distribution over a fixed vocabulary, conditioned on the source representations $\mathbf{h}_{enc}$ and previously generated tokens represented as $\mathbf{h}^{t-1}_{dec}$:
% : $p(t_1,\dots,t_{L_{\mathbf{t}}}|b_1,\dots,b_{L_{\mathbf{b}}})$.
    \begin{equation*}
        \label{eq:p_vocab}
        \begin{aligned}
            p_{g}(y^t|y^{1,...,t-1},\mathbf{x}) &= f_{out}(y^{t-1}, \mathbf{h}^t_{dec},\mathbf{c})
        \end{aligned}
    \end{equation*}
    where $f_{out}$ is a non-linear function, typically a softmax classifier with an attention mechanism, that outputs the probabilities over all the words in a preset vocabulary $\mathcal{V}$. 

\smallsection{Our Tailored Seq2Seq Generative Model}
We use guided beam search to generate diverse keyphrases for each document.
Previous work~\cite{Meng_2017} has shown that even when the gold labels are available, a vanilla Seq2Seq model would collapse and fail to generate high-quality candidate phrases. 
Since we only train the model with silver labels, to improve the generating quality, we encourage the decoder model to generate words that appear in the input document $\mathbf{x}$.
More specifically, we double the probabilities of the words occurred in the input document.
Note that, words which do not appear in the input document can still be generated so the diversity can be maintained. 
This also matches our observation that many absent keyprhases have all their tokens in the input document.

\smallsection{Relationship to Copy Mechanism}
In fact, our tailored Seq2Seq model reassembles the copy mechanism proposed in~\cite{Meng_2017} and can be viewed as a special version by assuming all tokens in the input documents follows a similar distribution as estimated by the encoder-decoder model.

As shown in~\citet{Meng_2017}, the copy mechanism is useful for generating keyword extraction because it gives high probabilities to the words that exist in the input document. 
This is achieved by an extra probability term.
\begin{equation*}
    \begin{aligned}
    p_c(y^t|y^{1,...,t-1},\mathbf{x})&=\frac{1}{Z}\mathop{\sum}_{j:x_j=y^t} exp(\psi(\mathbf{x}_j)),y^t \in \mathbf{x} \\
    \psi(\mathbf{x}_j) &= \sigma((\mathbf{h}^j_{dec})^T W)s^t,
    \end{aligned}
\end{equation*}
where $\sigma$ is a non-linear function, $W$ is a learned parameter matrix, and $Z$ is the sum of the scores used for normalization. 
%\jingbo{I think there are some terms not explained here. $\psi$? $p_c$?}
For CopyRNN, the probability of generating $y^t$ is the sum of $p_g$ and $p_c$.
%\jingbo{$p_g$ and $p_c$?}

% We have included a CopyRNN variant of \our in experiments for comparison.

% \smallsection{Generation and Reranking}
% For a given (unseen) input document, we employ beam search to generate several candidate phrases from the trained generative model.
% The generated keyphrases will be added to the candidate pool, and then we will follow the same ranking module to rank them again.

% However, the length of output is usually smaller than 5, so unigram will have much higher probability to be generated. 
% To solve this problem, we will only consider several top unigrams with the maximum generating probability.

\section{Experiments}
\label{sec:experiment}
In this section, we first introduce datasets used in this study, followed by baselines, evaluation metrics, and details of implementation. 
Then, we present and discuss the experiment results of present keyphrase and absent keyphrase generation.

\begin{table}[t]
    \centering
    \caption{Statistics of datasets. Only the supervised model CopyRNN uses document-keyphrase labels and the validation set. 
    All other methods use raw documents from the KP20k training set as input.} \label{tab:dataset_summary}
    \small
    \vspace{-3mm}
    \begin{tabular}{cccc}
        \toprule
        Dataset & Train & Valid & Test \\
        \midrule
        KP20k & 514,154 & 19,992 & 19,987 \\
        Inspec & - & 1,500 & 500 \\
        Krapivin & - & 1,844 & 460 \\
        NUS & - & - & 211 \\
        SemEval & - & 144 & 100 \\
        \bottomrule
    \end{tabular}
\end{table}

\begin{table*}[t]
  \centering
  \caption{F$_1$ scores of present keyphrase prediction on five scientific publication datasets. 
  ExpandRank is too slow to be evaluated on the KP20k dataset.
  Supervised-CopyRNN results are from its original work~\cite{Meng_2017}.}
  \vspace{-3mm}
  \scriptsize
   \renewcommand{\arraystretch}{1.2}
  \begin{tabular}{cccccccccccccccc}
    \toprule
    \multicolumn{1}{c}{}
    & \multicolumn{3}{c}{\textbf{Kp20K}}
    & \multicolumn{3}{c}{\textbf{Inspec}}
    & \multicolumn{3}{c}{\textbf{Krapivin}}
    & \multicolumn{3}{c}{\textbf{NUS}}
    & \multicolumn{3}{c}{\textbf{SemEval}}
    \\
    \midrule
    \textbf{Model} 
    & \fontsize{7}{8}\selectfont{\textbf{@5}} & \fontsize{7}{8}\selectfont{\textbf{@10}} & \fontsize{7}{8}\selectfont{\textbf{@$\mathcal{O}$}}\xspace 
    & \fontsize{7}{8}\selectfont{\textbf{@5}} & \fontsize{7}{8}\selectfont{\textbf{@10}} & \fontsize{7}{8}\selectfont{\textbf{@$\mathcal{O}$}}\xspace 
    & \fontsize{7}{8}\selectfont{\textbf{@5}} & \fontsize{7}{8}\selectfont{\textbf{@10}} & \fontsize{7}{8}\selectfont{\textbf{@$\mathcal{O}$}}\xspace
    & \fontsize{7}{8}\selectfont{\textbf{@5}} & \fontsize{7}{8}\selectfont{\textbf{@10}} & \fontsize{7}{8}\selectfont{\textbf{@$\mathcal{O}$}}\xspace 
    & \fontsize{7}{8}\selectfont{\textbf{@5}} & \fontsize{7}{8}\selectfont{\textbf{@10}} & \fontsize{7}{8}\selectfont{\textbf{@$\mathcal{O}$}}\xspace  \\
    \midrule
    TF-IDF & 7.2 & 9.4 & 6.3 & 24.2 & 28.0 & 24.8 & 11.5 & 14.0 & 13.3 & 11.6 & 14.2 & 12.5 & 16.1 & 16.7 &  15.3 \\
    SingleRank & 9.9 & 12.4 & 10.3 & 21.4 & 29.7 & 22.8 & 9.6 & 13.6 & 13.4 & 13.7 & 16.2 & 18.9 & 13.2 & 16.9 & 14.7  \\
    % \textbf{ExpandRank} & - & - & - & 21.1 & 29.5 & - & 9.6 & 13.6 & - & 13.7 & 16.2 & - & 13.5 & 16.3 & - \\
    TextRank & 18.1 & 15.1 & 14.1 & 26.3 & 27.9 & 26.0 & 14.8 & 13.9 & 13.0 & 18.7 & 19.5 & 19.9 & 16.8 & 18.3 & 18.1\\ 
    ExpandRank &N/A & N/A & N/A & 21.1 & 29.5 & 26.8 & 9.6 & 13.6 & 11.9&13.7 & 16.2 &15.7& 13.5 & 16.3&14.4 \\
    EmbedRank &15.5 &15.6 & 15.8 & 29.5 & 34.4 & 32.8 & 13.1 & 13.8 & 13.9&10.3 & 13.4 &14.7& 10.8 & 14.5&13.9 \\
    \our  & \textbf{23.4} & \textbf{24.6} & \textbf{23.8} & \textbf{30.3} & \textbf{34.5} & \textbf{33.1} & \textbf{17.1} & \textbf{15.5} & \textbf{15.8} & \textbf{21.8} & \textbf{23.3} & \textbf{23.7} & \textbf{18.7} & \textbf{24.0} & \textbf{22.7} \\
    \midrule
    \our-OnlyBank & 22.9 & 23.1 & 23.1 & 29.7 & 32.8 & 32.1 & 15.9 & 14.3 & 14.2 & 20.7 & 21.8 & 22.3 & 16.3 & 20.9 & 20.4\\ 
     \our-OnlyEmbed & 21.2&22.9&21.8& 29.7&34.8& 32.7 &15.9 &16.4& 14.3& 20.4&21.3&22.6 & 15.3&16.5&15.9  \\
    %\our-CopyRNN  &22.7 & \textbf{25.7}&\textbf{24.4} & 30.1 & \textbf{35.5} & \textbf{34.8} & 16.6 & \textbf{16.8} & \textbf{16.3} & \textbf{22.8} & \textbf{25.3} & \textbf{24.2} & 17.2 & \textbf{24.8} & \textbf{23.7} \\
    \midrule
    Supervised-CopyRNN & \textbf{32.8} & 25.5 &N/A & 29.2 & 33.6 &N/A & \textbf{30.2} & \textbf{25.2} & N/A & \textbf{34.2} &  \textbf{31.7} &  N/A & \textbf{29.1} & \textbf{29.6} & N/A \\
    \bottomrule
  \end{tabular}
  \label{tab:present_exp}
\end{table*}

\subsection{Datasets}
    
    % Due to the public availability, scientific publication dataset are widely used to evaluate keyphrase generation methods. 
    We follow previous keyphrase generation studies~\citep{Meng_2017,ye_18,meng2019does,chen2019guided} 
    % \jingbo{Some citations here} 
    and adopt five scientific publication datasets for evaluation.
    \textbf{KP20k} is the largest dataset in scientific keyphrase studies thus far.
    % , which provides train/dev/test splits. 
    % This dataset contains a large amount of high-quality scientific publications covering different computer science domains. 
    There are four other widely-used scientific datsets for comparing different models: \textbf{Inspec}~\cite{tomokiyo2003language}, \textbf{Krapivin}~\cite{Krapivin2009LargeDF}, \textbf{NUS}~\cite{nguyen2007keyphrase}, and \textbf{SemEval-2010}~\cite{kim2010semeval}.
    Table~\ref{tab:dataset_summary} presents the details of all datasets\footnote{Dataset release is from \url{https://github.com/memray/OpenNMT-kpg-release}}.
    
    All the models in our experiments are built on the KP20k training set. 
    Only the supervised model CopyRNN uses document-keyphrase labels and the validation set. 
    All other methods use raw documents from the KP20k training set as input.
    Once the model is built, it will be applied to all the five test sets for evaluations.
    
    % Because KP20k provides a much larger number of documents than other datasets, we train our tailored seq2seq generative model with silver data from the KP20k dataset training set and apply this trained model to all the five test sets. 
    
    %\jingbo{I think it's important to make it explicit that all the methods will be trained on the training set and tested on the test set. You may want to mention this is also following the literature. What's the role of the validation set? If it's not used, maybe say something here, e.g., validation set is only used for supervised methods.}

\subsection{Compared Methods}
    We compare \our with five other unsupervised methods. 
    \begin{itemize}[nosep,leftmargin=*]
        \item \textbf{TF-IDF}~\cite{Jones72astatistical} ranks the extracted noun phrase candidates by term frequency and inverse document frequency in the given documents.
        \item \textbf{TextRank}~\cite{mihalcea2004textrank} simulates the word as web page, then uses the PageRank algorithm to find the keyphrases.
        \item \textbf{ExpandRank}~\cite{florescu-caragea-2017-positionrank} is an extension of TextRank utilizing Emebedding similarity to get neighbouring documents to set a better edge weight in the PageRank~\citep{page1999pagerank} algorithm.
        \item \textbf{EmbedRank}~\cite{DBLP:conf/conll/Bennani-SmiresM18} directly uses embedding similarity to rank the present candidate keyphrase and uses Maximal Marginal Relevance (MMR)~\cite{DBLP:journals/sigir/CarbinellG17} to increase the diversity of extracted keyphrases.
    \end{itemize}
    
    \noindent For ablation studies, we compare some variants of our \textbf{\our} method as follows.
    \begin{itemize}[nosep,leftmargin=*]
        \item \textbf{\our-OnlyBank} only uses the partial match between the phrase bank and the input document to extract keyphrase candidates without any seq2seq model.
        \item \textbf{\our-OnlyEmbed} ranks the candidate phrases with only the embedding similarity without the TF-IDF information.
        % \item \textbf{\our-CopyRNN} utilizes the CopyRNN to replace our tailored seq2seq model. 
    \end{itemize}
    
    \noindent We also present \textbf{Supervised-CopyRNN}~\cite{Meng_2017}, which trains CopyRNN on the \emph{labeled} KP20K dataset to generate keyphrases.
    Since it is trained based on gold labels, we regard it as an upper bound of all other unsupervised methods.

\begin{table*}[t]
  \centering
  \caption{Recall scores of absent keyphrase prediction on five scientific publications datasets. ExpandRank is too slow to be evaluated on the KP20k dataset.} 
  \vspace{-3mm}
  \small
  \begin{tabular}{ccccccccccc}
    \toprule
    \multicolumn{1}{c}{}
    & \multicolumn{2}{c}{\textbf{Kp20K}}
    & \multicolumn{2}{c}{\textbf{Inspec}}
    & \multicolumn{2}{c}{\textbf{Krapivin}}
    & \multicolumn{2}{c}{\textbf{NUS}}
    & \multicolumn{2}{c}{\textbf{SemEval}}
    \\
    \midrule
    Model 
    & \fontsize{7}{8}\selectfont{\textbf{R@10}} & \fontsize{7}{8}\selectfont{\textbf{R@20}} 
    & \fontsize{7}{8}\selectfont{\textbf{R@10}} & \fontsize{7}{8}\selectfont{\textbf{R@20}}  
    & \fontsize{7}{8}\selectfont{\textbf{R@10}} & \fontsize{7}{8}\selectfont{\textbf{R@20}}  
    & \fontsize{7}{8}\selectfont{\textbf{R@10}} & \fontsize{7}{8}\selectfont{\textbf{R@20}}  
    & \fontsize{7}{8}\selectfont{\textbf{R@10}} & \fontsize{7}{8}\selectfont{\textbf{R@20}}  \\
    \midrule
    Other Unsupervised Methods & 0 & 0 & 0 & 0 & 0 & 0 & 0 & 0 & 0 & 0 \\
    ExpandRank & N/A & N/A & 0.02 & 0.05 & 0.01 & 0.015 & 0.005 & 0.04 & 0 & 0.004 \\
    \our & \textbf{2.3} & \textbf{2.5} & \textbf{1.7} & \textbf{2.1} & \textbf{3.3} & \textbf{5.4} & \textbf{2.4} & \textbf{3.2} & \textbf{1.0} & \textbf{1.1} \\
    \midrule
    \our-OnlyBank & 1.8 & 2.2 & 1.5 & 1.7 & 3.1 & 4.1 & 2.1 & 2.6 & 0.7 & 0.9\\
    % \our-CopyRNN &1.8 &2.0 & 1.6 & 1.9 & 3.1 & 4.7 & 1.9 &2.8 & 1.0 & 1.1\\
    \midrule
     Supervised-CopyRNN & \textbf{11.5}  & \textbf{14.0} & \textbf{5.1} & \textbf{6.8} & \textbf{11.6} & \textbf{14.2} & \textbf{7.8} & \textbf{10.0} & \textbf{4.9} & \textbf{5.7} \\
    \bottomrule
  \end{tabular}
  \label{tab:absent_exp}
\end{table*}

\subsection{Evaluation Metrics}\label{subsec:metric}
% \jingbo{I think you may want to introduce $F_1@K$ and $F_1@O$ a bit. These two are not very popular, especially the latter one.} 
    Following the literature, we evaluate the model performance on generating present and absent keyphrases separately. 
    If some models generate the two types of keyphrases in a unified ranked list, we split them into two ranked lists by checking whether or not the phrases appear in the input document. 
    The relative ranking between the phrases of the same type is therefore preserved.
    % Both lists are sorted with score $S$ being the key and returned as present keyphrase ranked list and absent keyphrase ranked list. 

    We use $R@k$, $F_1@k$, and $F_1@\mathcal{O}$~\cite{yuan2018size} as main evaluation metrics.
    Specifically, $F_1@5$, $F_1@10$, and $F_1@\mathcal{O}$ are utilized for evaluating present keyphrases and $R@10$ and $R@20$ for absent keyphrases.
    We report the macro-average scores over all documents in each test set.
    
    Specifically, given a ranked list of keyphrases, either present or absent, $\mathcal{\hat{Y}}=(\mathbf{\hat{y}}_1, \ldots, \mathbf{\hat{y}}_{|\mathcal{\hat{Y}}|})$ and the corresponding groundtruth keyphrase set $\mathcal{Y}$, we first truncate it with a cutoff $k$ (i.e., $\mathcal{\hat{Y}}_{:k}=(\mathbf{\hat{y}}_1, \ldots, \mathbf{\hat{y}}_{\min{(k, |\mathcal{\hat{Y}}|  )}})$) and then evaluate its precision and recall:
    \begin{equation*}
      P@k = \frac{|\mathcal{\hat{Y}}_{:k} \cap \mathcal{Y}|}{|\mathcal{\hat{Y}}_{:k}|},  R@k = \frac{|\mathcal{\hat{Y}}_{:k} \cap \mathcal{Y}|}{|\mathcal{Y}|}
    \end{equation*}
    $F_1@k$ is the harmonic mean of $P@k$ and $R@k$.
    $F_1@\mathcal{O}$ can be viewed as a special case of $F_1@k$ when $k=|\mathcal{Y}|$.
    In other words, we only examine the same amount of keyphrases as the number of our groundtruth keyphrases.
    
    We apply Porter Stemmer provided by NLTK~\citep{bird2009natural} to both ground-truth and predicted keyphrases to determine whether phrases appear in the original document and whether two keyphrases match or not.
 
\subsection{Implementation Details}
\label{subsec:implementation}
    
    For all the methods that involve keyphrase extraction, we utilize the open-source toolkit pke\footnote{\url{https://github.com/boudinfl/pke}} for phrase candidate generation.
    The window size of the graph-based models SingleRank, TextRank and ExpandRank has been searched from 2 to 10, and again, the best performance is selected.

    The vocabulary $\mathcal{V}$ in the seq2seq model consists of 50,000 most frequent words.  
    We train the model for 500,000 steps and select the last checkpoint for inference. 
    The dimension of LSTM cell is 256, the embedding dimension is 200, and the max length of source text is 512.
    Models are optimized using Adagrad~\citep{duchi2011adaptive} with initial learning rate sets to $0.001$, and will be linearly decayed by 0.8 after every 5 epochs. 
    The beam size for keyphrase generating beam search is 20. 

\subsection{Present Keyphrase Evaluation}
\label{subsec:present}
    The results of present keyphrase generation are listed in Table~\ref{tab:present_exp}.
    % Overall, \our and \our-CopyRNN achieves the best $F_1$@5, $F_1$@10 and $F_1$@$\mathcal{O}$ performances among all the unsupervised methods.
    Overall, \our achieves the best $F_1$@5, $F_1$@10 and $F_1$@$\mathcal{O}$ performances among all the unsupervised methods. 
    EmbedRank is arguably the strongest baseline method, however, \our outperforms it on many datasets with a significant margin. 
    % Only present keyphrases are considered. 
    % The ExpandRank baseline fails to return any result on the KP20k dataset, due to its high time complexity. 
    
    % We observe that \our-CopyRNN performs better on the $F1@10$  and $F_1@\mathcal{O}$ scores on all five datasets, but there is stable improvement for the $F1@5$ score. 
    
    %\jingbo{Could you please double check the numbers in the Table? It seems to me this is not the case anymore. Not sure if you have updated any results?}
    % It shows that our tailored seq2seq generative model can produce a larger number of reliable keyphrases than using CopyRNN. 
    % Their difference will be more obvious in the absent keyphrase comparison.

    One can easily see that \our outperforms on all the datasets than \our-OnlyEmbed.
    It shows that the TF-IDF information adds values to the embedding-based ranking heuristic.
    The \our-OnlyEmbed model performs about the same as \our on the Inspec dataset, because the length of document in the Inspec dataset is the shortest among all other dataset.
    As we discussed earlier, TF-IDF is more stable when the documents are sufficiently long.
    
    The obvious advantage of \our over \our-OnlyBank demonstrates that our generation module does generate some ``novel'' present phrases beyond the scope of the extractor. 

    It is worth mentioning that on the Inspec dataset, \our is even better than the Supervised-CopyRNN method.

\begin{figure*}[ht]
    \centering
      \includegraphics[width=1.0\textwidth]{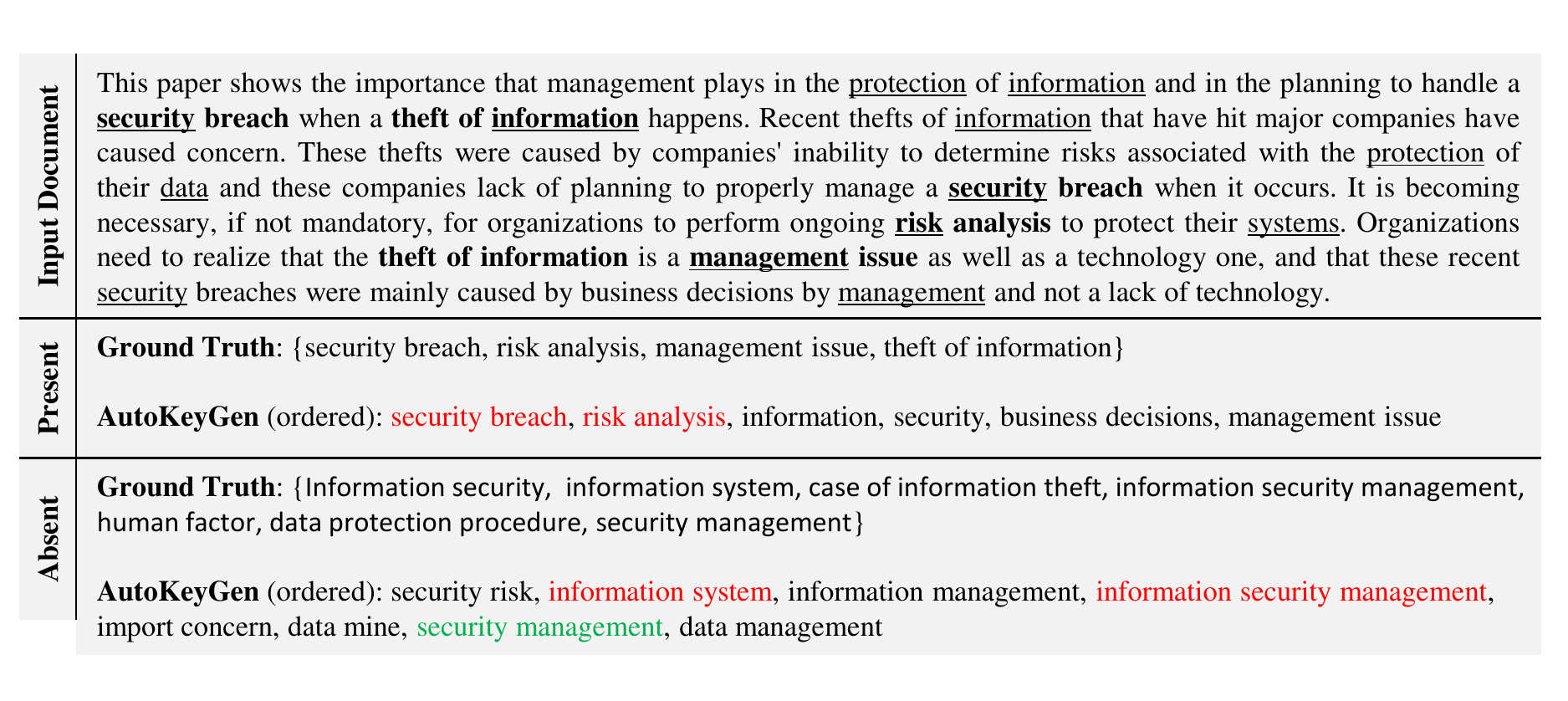}
      
    \vspace{-3mm}
    \caption{A case study of \our from the NUS test set. Present keyphrases are marked bold in the input document. Tokens in the input document related to absent keyphrases are underlined. 
    Correctly predicted keyphrases are highlighted in red.
    The green one is a correct phrase predicted by our generating module, which is omitted by noun phrase extraction method.}  %\jingbo{This case study is not very interesting. Can we find some cases that there is one absent keyphrase is actually a present keyphrase in another document? And ideally, this document also has another absent keyphrases generated by our RNN model. If it's hard to find such a document, we can split the two types of successful absent keyphrase generation into two documents. Try to tell a story supporting our observations and method, instead of merely presenting the results.}}
    \label{fig:case}
\end{figure*}

\subsection{Absent Keyphrase Evaluation}
\label{subsec:absent}

    Table~\ref{tab:absent_exp} presents the model comparison on absent keyphrase generation.
    Following ~\cite{Meng_2017}, only recall score is reported as comparison.
    Since all unsupervised baseline methods except ExpandRank are not capable of generating any absent keyphrases, we refer to them together as ``\emph{Other Unsupervised Methods}''.
    Among all unsupervised models, \our has the best recall on all the datasets. 
    Therefore, we argue that \our unleashes the potential to derive high-quality absent keyphrases under the unsupervised setting.
    
    Comparing \our with \our-OnlyBank, one can tell that the generation module does help improve the performance.
    
    % We notice that the performance of \our-CopyRNN is similar to \our-OnlyBank.
    % This shows CopyRNN is not as effective as our tailored seq2seq when silver data is used for training.
    % We conjecture that this is caused by the copy mechanism used in the CopyRNN, which encourages to generate frequent tokens in the input document.
    % Such tokens may have been already included in our absent candidates from the phrase bank. 

\begin{figure}[t]
 \centering
 \includegraphics[width=0.8\linewidth]{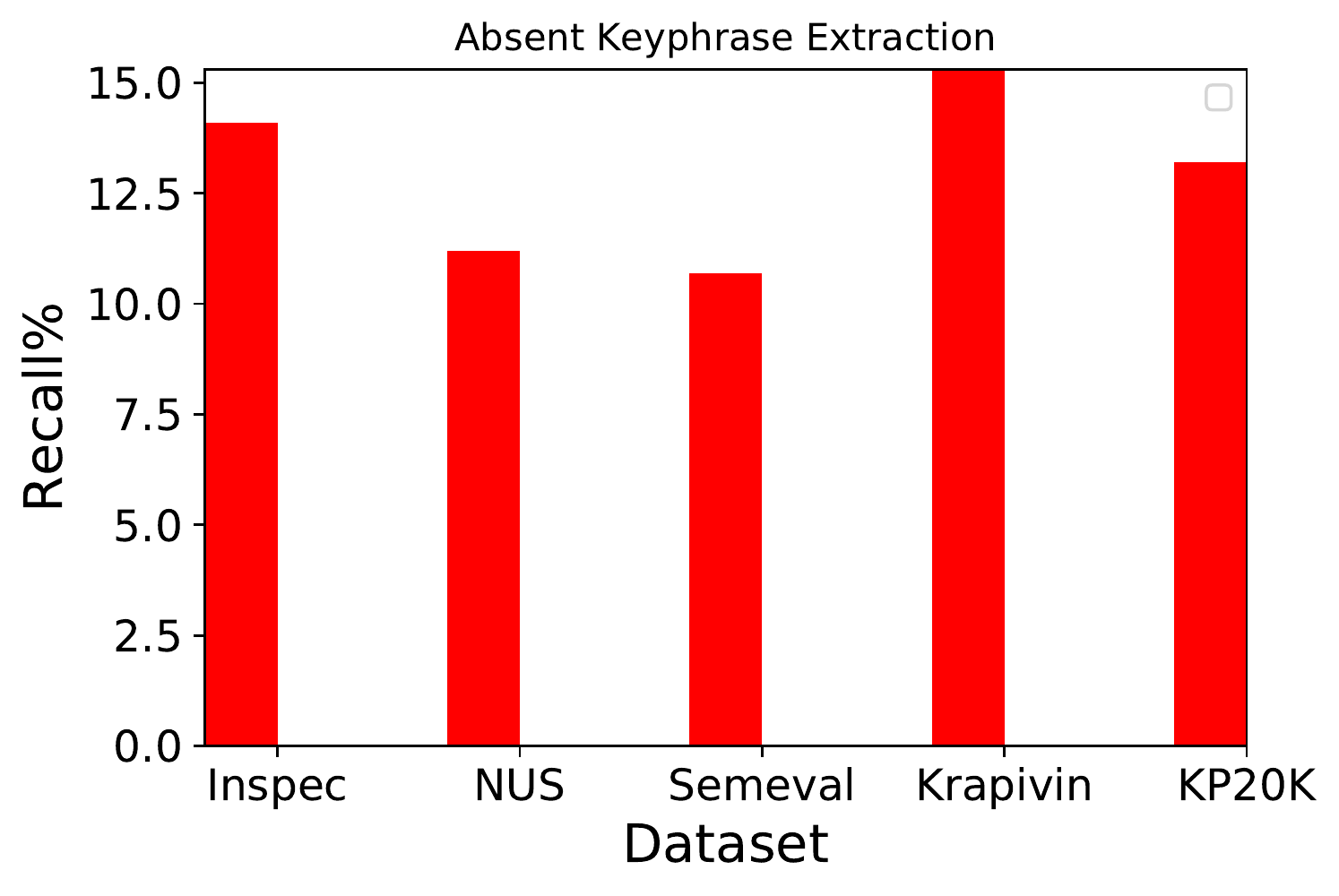}
   
    \vspace{-3mm}
    \caption{The recall of absent keyphrases using all the phrases in phrase bank on five datasets.}
 \label{fig:absent}
    \vspace{-3mm}
\end{figure}

\subsection{Case Studies}
Figure~\ref{fig:case} presents a case study from the NUS test set. 
Parts of this case study have been presented in the overview of \our, i.e., Figure~\ref{fig:overview}.

``\emph{Recent security breach}'' is extracted as a keyphrase by the conventional noun phrase extractor, but our method successfully removes the ``recent'' and generates the phrase candidate ``\emph{security branch}'' which is a groundtruth present keyphrase. 
This is mainly benefited from our tailored extractor.

As for the absent keyphrase, our method successfully generated ``\emph{information system}'' and ``\emph{information security management}'' from the phrase bank.
That is, these two phrases were extract from other documents. 
Since all their tokens appear in this document, they are added as absent candidates.

Our method can not only obtain absent keyphrases from the phrase bank, but also generate keyphrases from the tailored seq2seq generative model.
In this case, ``\emph{security risk}'', ``\emph{information management}'', and ``\emph{security management}'' are all generated by the generation module.
Although some of them are not perfectly matched with absent ground truth keyphrases, they contain similar meanings.
Therefore, we believe our generative model does have a potential to produce reliable absent keyphrases.

\subsection{Candidate Absent Keyphrase Quality}

%\jingbo{I don't think we want to compare our tailored one vs. the standard one anymore. Let's just present the recall for the present keyphrases from the extraction. Perhaps we can merge Figure 2 and Figure 3 together? Dual Y-Axis could be a choice if it looks beautiful.}
%all the extracted keyphrases by the standard noun phrase extraction method and our proposed noun phrase extraction method. We can observe our method is able to achieve a higher recall of present keyphrase.

%\jingbo{I'm not sure why this is an upperbound? We do have a generative model for phrases beyond the phrase bank, right?}
Figure~\ref{fig:absent} presents the intersection between the phrase bankd and the groundtruth absent keyphrase. 
%\jingbo{do you mean that the intersection between the phrase bank and all the groundtruth absent keyphrases? If so, please make it clear.}
It serves as an upper bound of the recall of the absent keyphrases for the extractive part of \our.
However, such upper bounds are very loose, as the number of generated absent candidates from the phrase bank is too big.
% It is because we are unable to select all the noun phrases from the dataset as the absent candidate phrases, considering the size of candidate phrase pool is too big. Therefore, we employ a relatively strict method to filter all the absent candidate phrases.  
However, this does suggest that there is a great potential of the deep unsupervised keyphrase generation, if one can come up with a better ranking module for absent keyphrases.

\section{Related Work}
\label{sec:related_work}

In this section, we mainly review the literature related to the following three areas, (1) keyphrase generation, (2) word and document embeddings, and (3) encoder-decoder models. 

\subsection{Kerphrase Generation}

Most of the existing algorithms have addressed the task of keyphrase extraction through two steps~\citep{liu-etal-2009-clustering,tomokiyo2003language}. 
The first step is to acquire a list of keyphrase candidates. 
Previous studies use n-grams or noun phrases with certain part-of-speech patterns to identify potential candidates~\citep{10.3115/1119355.1119383, le2016unsupervised, wang2016ptr}. AutoPhrase~\citep{Shang_2018} serves as another option to extract high-quality candidates, using a distant supervised phrase mining method leveraging open-domain knowledge, such as Wikipedia.
The second step is to rank candidates on their importance to the document using either supervised or unsupervised approaches with manually-defined features~\citep{ 10.5555/1642293.1642576,  florescu-caragea-2017-positionrank}. ~\citet{florescu-caragea-2017-positionrank} tries to score the candidate phrases as the aggregation of its words score, but over-generation erros will happen.
~\citet{DBLP:conf/coling/SaxenaMJ20} transforms keyphrase extraction into classification problem using evolutionary game theory.

The major common drawback of these keyphrase extraction methods is that they can only extract keyphrases that already appear in the source text and thus they fail to predict keyphrases in a different word order or some synonymous keyphrases. 

To address this issue, keyphrase generation methods have been proposed such as CopyRNN~\cite{Meng_2017} and CopyCNN~\cite{8248519}. 
These methods utilize an encoder-decoder architecture, treating the title and main text body as the source information and keyphrases as the target to predict. 
However, those approaches ignore the leading role of the title in the document structure. 
To fully leverage the title information, \citet{ye_18} proposed a semi-supervised learning approach that generates more training pairs and \citet{chen2019guided} proposed to take title features as a query to guide the decoding process.
\citet{DBLP:conf/emnlp/SwaminathanZMGS20} firstly applies GAN to keyphrase extraction problem, and it presents a new promising direction for keyphrase extraction problem.

Our work is fully unsupervised, thus being significantly different from these existing generation methods that rely on human annotations.

\subsection{Word and Document Embeddings}
Embddings~\cite{41224} represents words as vectors in a continuous vector space. It's widely used in many NLP problems, since embeddings methods take advantages over the classic bag-of-words representation considering it can capture semantic relatedness with acceptable dimensions.
The state-of-the-art embeddings methods such as ~\cite{DBLP:conf/rep4nlp/LauB16} is able to infer a vector of a document via a embedding network. In this way, the embeddings of a short phrase and a long document can be represented in a shared vector space, which make it feasible to derive their semantic relatedness directly with the embedding similarity.

\subsection{Encoder-Decoder Model}
The RNN-based encoder-decoder architecture was first introduced by ~\citet{cho-etal-2014-learning} and ~\citet{sutskever2014seq2seq} for machine translation problems. It has also achieved great successes in many other NLP tasks~\cite{DBLP:conf/aaai/SerbanSBCP16,DBLP:journals/ile/LiuSA19}. 
Encoder-decoder model is also used for keyphrase extraction problem. Some work~\citep{DBLP:conf/acl/ChenCLK20, DBLP:journals/corr/AllamanisPS16}  tried to copy certain parts of source text when generating the output.  
~\citet{See_2017} enhanced this architecture with a pointer-generator network, which allows models to copy words from the source text. ~\citet{celikyilmaz-etal-2018-deep} proposed an abstractive system where multiple encoders represent the document together with a hierarchical attention mechanism for decoding.

\section{Conclusions and Future Work}
In this paper we propose an unsupervised deep keyphrase generation method to derive present keyphrases and absent keyphrases from the document itself.
% We combine retrieve method and generate method together, and rank them with a heuristic function.
Our design is inspired by two intuitive observations.
Extensive experiments demonstrate the superiority of our method against existing unsupervised models in terms of both present and absent keyphrases. 
% Our method overcomes the ineffectiveness of embeddings-based method~\cite{DBLP:conf/conll/Bennani-SmiresM18} for long documents. 

% There are still some limitations of our work. 
% First, we need a better constraint to obtain reliable absent candidate phrases.
In the future, we plan to enhance the silver label quality for the deep generative model, so the absent keyphrase generation could be further improved. 
One possible way is to filter the candidate phrases according to the keyphrases correlations. 
Another promising direction is to leverage the intrinsic article structure, such as title-body relations, for a self-supervised learning.

\section*{Acknowledgements}
Our work is supported in part by NSF Convergence Accelerator under award OIA-2040727. Any opinions, findings, and conclusions or recommendations expressed herein are those of the authors and should not be interpreted as necessarily representing the views, either expressed or implied, of the U.S. Government. The U.S. Government is authorized to reproduce and distribute reprints for government purposes notwithstanding any copyright annotation hereon.
% \section{Ethical Considerations}
% We do not anticipate any significant ethical concern.
% The datasets used in our experiments are available in previous related works, they are open resources on Internet; our work can be applied to search engine, or other information retrieval system. they seem to be low-risk applications for us; `attributing identity characteristics' is avoided in our work; our work doesn't require much computing resources, every experiments can be finished on a laptop with GPU.

\bibliographystyle{acl_natbib}
\bibliography{naacl2021}

\begin{thebibliography}{48}
\expandafter\ifx\csname natexlab\endcsname\relax\def\natexlab#1{#1}\fi

\bibitem[{Allamanis et~al.(2016)Allamanis, Peng, and
  Sutton}]{DBLP:journals/corr/AllamanisPS16}
Miltiadis Allamanis, Hao Peng, and Charles Sutton. 2016.
\newblock \href {http://arxiv.org/abs/1602.03001} {A convolutional attention
  network for extreme summarization of source code}.
\newblock \emph{CoRR}, abs/1602.03001.

\bibitem[{Alzaidy et~al.(2019)Alzaidy, Caragea, and Giles}]{alzaidy2019bi}
Rabah Alzaidy, Cornelia Caragea, and C~Lee Giles. 2019.
\newblock Bi-lstm-crf sequence labeling for keyphrase extraction from scholarly
  documents.
\newblock In \emph{The world wide web conference}, pages 2551--2557.

\bibitem[{Bennani{-}Smires et~al.(2018)Bennani{-}Smires, Musat, Hossmann,
  Baeriswyl, and Jaggi}]{DBLP:conf/conll/Bennani-SmiresM18}
Kamil Bennani{-}Smires, Claudiu Musat, Andreea Hossmann, Michael Baeriswyl, and
  Martin Jaggi. 2018.
\newblock \href {https://doi.org/10.18653/v1/k18-1022} {Simple unsupervised
  keyphrase extraction using sentence embeddings}.
\newblock In \emph{Proceedings of the 22nd Conference on Computational Natural
  Language Learning, CoNLL 2018, Brussels, Belgium, October 31 - November 1,
  2018}, pages 221--229. Association for Computational Linguistics.

\bibitem[{Bird et~al.(2009)Bird, Klein, and Loper}]{bird2009natural}
Steven Bird, Ewan Klein, and Edward Loper. 2009.
\newblock \emph{Natural language processing with Python: analyzing text with
  the natural language toolkit}.
\newblock " O'Reilly Media, Inc.".

\bibitem[{Boudin and Gallina(2021)}]{boudin2021redefining}
Florian Boudin and Ygor Gallina. 2021.
\newblock Redefining absent keyphrases and their effect on retrieval
  effectiveness.
\newblock \emph{arXiv preprint arXiv:2103.12440}.

\bibitem[{Campos et~al.(2018)Campos, Mangaravite, Pasquali, Jorge, Nunes, and
  Jatowt}]{DBLP:conf/ecir/0001MPJNJ18}
Ricardo Campos, V{\'{\i}}tor Mangaravite, Arian Pasquali,
  Al{\'{\i}}pio~M{\'{a}}rio Jorge, C{\'{e}}lia Nunes, and Adam Jatowt. 2018.
\newblock \href {https://doi.org/10.1007/978-3-319-76941-7\_63} {A text feature
  based automatic keyword extraction method for single documents}.
\newblock In \emph{Advances in Information Retrieval - 40th European Conference
  on {IR} Research, {ECIR} 2018, Grenoble, France, March 26-29, 2018,
  Proceedings}, volume 10772 of \emph{Lecture Notes in Computer Science}, pages
  684--691. Springer.

\bibitem[{Carbinell and Goldstein(2017)}]{DBLP:journals/sigir/CarbinellG17}
Jaime Carbinell and Jade Goldstein. 2017.
\newblock \href {https://doi.org/10.1145/3130348.3130369} {The use of mmr,
  diversity-based reranking for reordering documents and producing summaries}.
\newblock \emph{{SIGIR} Forum}, 51(2):209--210.

\bibitem[{Celikyilmaz et~al.(2018)Celikyilmaz, Bosselut, He, and
  Choi}]{celikyilmaz-etal-2018-deep}
Asli Celikyilmaz, Antoine Bosselut, Xiaodong He, and Yejin Choi. 2018.
\newblock \href {https://doi.org/10.18653/v1/N18-1150} {Deep communicating
  agents for abstractive summarization}.
\newblock In \emph{Proceedings of the 2018 Conference of the North {A}merican
  Chapter of the Association for Computational Linguistics: Human Language
  Technologies, Volume 1 (Long Papers)}, pages 1662--1675, New Orleans,
  Louisiana. Association for Computational Linguistics.

\bibitem[{Chen et~al.(2020)Chen, Chan, Li, and King}]{DBLP:conf/acl/ChenCLK20}
Wang Chen, Hou~Pong Chan, Piji Li, and Irwin King. 2020.
\newblock \href {https://www.aclweb.org/anthology/2020.acl-main.103/}
  {Exclusive hierarchical decoding for deep keyphrase generation}.
\newblock In \emph{Proceedings of the 58th Annual Meeting of the Association
  for Computational Linguistics, {ACL} 2020, Online, July 5-10, 2020}, pages
  1095--1105. Association for Computational Linguistics.

\bibitem[{Chen et~al.(2019)Chen, Gao, Zhang, King, and Lyu}]{chen2019guided}
Wang Chen, Yifan Gao, Jiani Zhang, Irwin King, and Michael~R Lyu. 2019.
\newblock Title-guided encoding for keyphrase generation.
\newblock In \emph{Proceedings of the AAAI Conference on Artificial
  Intelligence}, volume~33, pages 6268--6275.

\bibitem[{Cho et~al.(2014)Cho, van Merri{\"e}nboer, Gulcehre, Bahdanau,
  Bougares, Schwenk, and Bengio}]{cho-etal-2014-learning}
Kyunghyun Cho, Bart van Merri{\"e}nboer, Caglar Gulcehre, Dzmitry Bahdanau,
  Fethi Bougares, Holger Schwenk, and Yoshua Bengio. 2014.
\newblock \href {https://doi.org/10.3115/v1/D14-1179} {Learning phrase
  representations using {RNN} encoder{--}decoder for statistical machine
  translation}.
\newblock In \emph{Proceedings of the 2014 Conference on Empirical Methods in
  Natural Language Processing ({EMNLP})}, pages 1724--1734, Doha, Qatar.
  Association for Computational Linguistics.

\bibitem[{Duchi et~al.(2011)Duchi, Hazan, and Singer}]{duchi2011adaptive}
John Duchi, Elad Hazan, and Yoram Singer. 2011.
\newblock Adaptive subgradient methods for online learning and stochastic
  optimization.
\newblock \emph{Journal of machine learning research}, 12(Jul):2121--2159.

\bibitem[{Florescu and Caragea(2017)}]{florescu-caragea-2017-positionrank}
Corina Florescu and Cornelia Caragea. 2017.
\newblock \href {https://doi.org/10.18653/v1/P17-1102} {{P}osition{R}ank: An
  unsupervised approach to keyphrase extraction from scholarly documents}.
\newblock In \emph{Proceedings of the 55th Annual Meeting of the Association
  for Computational Linguistics (Volume 1: Long Papers)}, pages 1105--1115,
  Vancouver, Canada. Association for Computational Linguistics.

\bibitem[{{Gers} and {Schmidhuber}(2001)}]{963769}
F.~A. {Gers} and E.~{Schmidhuber}. 2001.
\newblock \href {https://doi.org/10.1109/72.963769} {Lstm recurrent networks
  learn simple context-free and context-sensitive languages}.
\newblock \emph{IEEE Transactions on Neural Networks}, 12(6):1333--1340.

\bibitem[{Hasan and Ng(2010)}]{hasan2010conundrums}
Kazi~Saidul Hasan and Vincent Ng. 2010.
\newblock Conundrums in unsupervised keyphrase extraction: making sense of the
  state-of-the-art.
\newblock In \emph{Proceedings of the 23rd International Conference on
  Computational Linguistics: Posters}, pages 365--373. Association for
  Computational Linguistics.

\bibitem[{Hulth(2003)}]{10.3115/1119355.1119383}
Anette Hulth. 2003.
\newblock \href {https://doi.org/10.3115/1119355.1119383} {Improved automatic
  keyword extraction given more linguistic knowledge}.
\newblock In \emph{Proceedings of the 2003 Conference on Empirical Methods in
  Natural Language Processing}, EMNLP ’03, page 216–223, USA. Association
  for Computational Linguistics.

\bibitem[{Hulth and Megyesi(2006)}]{hulth-megyesi-2006-study}
Anette Hulth and Be{\'a}ta~B. Megyesi. 2006.
\newblock \href {https://doi.org/10.3115/1220175.1220243} {A study on
  automatically extracted keywords in text categorization}.
\newblock In \emph{Proceedings of the 21st International Conference on
  Computational Linguistics and 44th Annual Meeting of the Association for
  Computational Linguistics}, pages 537--544, Sydney, Australia. Association
  for Computational Linguistics.

\bibitem[{Jones(1972)}]{Jones72astatistical}
Karen~Spärck Jones. 1972.
\newblock A statistical interpretation of term specificity and its application
  in retrieval.
\newblock \emph{Journal of Documentation}, 28:11--21.

\bibitem[{Jones and Staveley(1999)}]{Jones1999PhrasierAS}
Steve Jones and Mark~S. Staveley. 1999.
\newblock Phrasier: a system for interactive document retrieval using
  keyphrases.
\newblock In \emph{SIGIR '99}.

\bibitem[{Kelleher and Luz(2005)}]{10.5555/1642293.1642576}
Daniel Kelleher and Saturnino Luz. 2005.
\newblock Automatic hypertext keyphrase detection.
\newblock In \emph{Proceedings of the 19th International Joint Conference on
  Artificial Intelligence}, IJCAI’05, page 1608–1609, San Francisco, CA,
  USA. Morgan Kaufmann Publishers Inc.

\bibitem[{Kim et~al.(2010)Kim, Medelyan, Kan, and Baldwin}]{kim2010semeval}
Su~Nam Kim, Olena Medelyan, Min-Yen Kan, and Timothy Baldwin. 2010.
\newblock Semeval-2010 task 5: Automatic keyphrase extraction from scientific
  articles.
\newblock In \emph{Proceedings of the 5th International Workshop on Semantic
  Evaluation}, pages 21--26.

\bibitem[{Krapivin et~al.(2009)Krapivin, Autaeu, and
  Marchese}]{Krapivin2009LargeDF}
Mikalai Krapivin, Aliaksandr Autaeu, and Maurizio Marchese. 2009.
\newblock Large dataset for keyphrases extraction.

\bibitem[{Lau and Baldwin(2016)}]{DBLP:conf/rep4nlp/LauB16}
Jey~Han Lau and Timothy Baldwin. 2016.
\newblock \href {https://doi.org/10.18653/v1/W16-1609} {An empirical evaluation
  of doc2vec with practical insights into document embedding generation}.
\newblock In \emph{Proceedings of the 1st Workshop on Representation Learning
  for NLP, Rep4NLP@ACL 2016, Berlin, Germany, August 11, 2016}, pages 78--86.
  Association for Computational Linguistics.

\bibitem[{Le et~al.(2016)Le, Nguyen, and Shimazu}]{le2016unsupervised}
Tho Thi~Ngoc Le, Minh~Le Nguyen, and Akira Shimazu. 2016.
\newblock Unsupervised keyphrase extraction: Introducing new kinds of words to
  keyphrases.
\newblock \emph{29th Australasian Joint Conference, Hobart, TAS, Australia,
  December 5-8, 2016}.

\bibitem[{Liu et~al.(2019)Liu, Sands{-}Meyer, and
  Audran}]{DBLP:journals/ile/LiuSA19}
Chenchen Liu, Sarah Sands{-}Meyer, and Jacques Audran. 2019.
\newblock \href {https://doi.org/10.1080/10494820.2018.1528283} {The
  effectiveness of the student response system {(SRS)} in english grammar
  learning in a flipped english as a foreign language {(EFL)} class}.
\newblock \emph{Interact. Learn. Environ.}, 27(8):1178--1191.

\bibitem[{Liu et~al.(2009)Liu, Li, Zheng, and Sun}]{liu-etal-2009-clustering}
Zhiyuan Liu, Peng Li, Yabin Zheng, and Maosong Sun. 2009.
\newblock \href {https://www.aclweb.org/anthology/D09-1027} {Clustering to find
  exemplar terms for keyphrase extraction}.
\newblock In \emph{Proceedings of the 2009 Conference on Empirical Methods in
  Natural Language Processing}, pages 257--266, Singapore. Association for
  Computational Linguistics.

\bibitem[{Meng et~al.(2019)Meng, Yuan, Wang, Brusilovsky, Trischler, and
  He}]{meng2019does}
Rui Meng, Xingdi Yuan, Tong Wang, Peter Brusilovsky, Adam Trischler, and Daqing
  He. 2019.
\newblock \href {http://arxiv.org/abs/1909.03590} {Does order matter? an
  empirical study on generating multiple keyphrases as a sequence}.

\bibitem[{Meng et~al.(2020)Meng, Yuan, Wang, Zhao, Trischler, and
  He}]{meng2020empirical}
Rui Meng, Xingdi Yuan, Tong Wang, Sanqiang Zhao, Adam Trischler, and Daqing He.
  2020.
\newblock An empirical study on neural keyphrase generation.
\newblock \emph{arXiv preprint arXiv:2009.10229}.

\bibitem[{Meng et~al.(2017)Meng, Zhao, Han, He, Brusilovsky, and
  Chi}]{Meng_2017}
Rui Meng, Sanqiang Zhao, Shuguang Han, Daqing He, Peter Brusilovsky, and
  Yu~Chi. 2017.
\newblock \href {https://doi.org/10.18653/v1/p17-1054} {Deep keyphrase
  generation}.
\newblock \emph{Proceedings of the 55th Annual Meeting of the Association for
  Computational Linguistics (Volume 1: Long Papers)}.

\bibitem[{Mihalcea and Tarau(2004)}]{mihalcea2004textrank}
Rada Mihalcea and Paul Tarau. 2004.
\newblock Textrank: Bringing order into text.
\newblock In \emph{Proceedings of the 2004 conference on empirical methods in
  natural language processing}, pages 404--411.

\bibitem[{Mikolov et~al.(2013)Mikolov, Chen, Corrado, and Dean}]{41224}
Tomas Mikolov, Kai Chen, Greg~S. Corrado, and Jeffrey Dean. 2013.
\newblock \href {http://arxiv.org/abs/1301.3781} {Efficient estimation of word
  representations in vector space}.

\bibitem[{Nguyen and Kan(2007)}]{nguyen2007keyphrase}
Thuy~Dung Nguyen and Min-Yen Kan. 2007.
\newblock Keyphrase extraction in scientific publications.
\newblock In \emph{International conference on Asian digital libraries}, pages
  317--326. Springer.

\bibitem[{Page et~al.(1999)Page, Brin, Motwani, and
  Winograd}]{page1999pagerank}
Lawrence Page, Sergey Brin, Rajeev Motwani, and Terry Winograd. 1999.
\newblock The pagerank citation ranking: Bringing order to the web.
\newblock Technical report, Stanford InfoLab.

\bibitem[{Saxena et~al.(2020)Saxena, Mangal, and
  Jain}]{DBLP:conf/coling/SaxenaMJ20}
Arnav Saxena, Mudit Mangal, and Goonjan Jain. 2020.
\newblock \href {https://doi.org/10.18653/v1/2020.coling-main.184} {Keygames:
  {A} game theoretic approach to automatic keyphrase extraction}.
\newblock In \emph{Proceedings of the 28th International Conference on
  Computational Linguistics, {COLING} 2020, Barcelona, Spain (Online), December
  8-13, 2020}, pages 2037--2048. International Committee on Computational
  Linguistics.

\bibitem[{See et~al.(2017)See, Liu, and Manning}]{See_2017}
Abigail See, Peter~J. Liu, and Christopher~D. Manning. 2017.
\newblock \href {https://doi.org/10.18653/v1/p17-1099} {Get to the point:
  Summarization with pointer-generator networks}.
\newblock \emph{Proceedings of the 55th Annual Meeting of the Association for
  Computational Linguistics (Volume 1: Long Papers)}.

\bibitem[{Serban et~al.(2016)Serban, Sordoni, Bengio, Courville, and
  Pineau}]{DBLP:conf/aaai/SerbanSBCP16}
Iulian~Vlad Serban, Alessandro Sordoni, Yoshua Bengio, Aaron~C. Courville, and
  Joelle Pineau. 2016.
\newblock \href
  {http://www.aaai.org/ocs/index.php/AAAI/AAAI16/paper/view/11957} {Building
  end-to-end dialogue systems using generative hierarchical neural network
  models}.
\newblock In \emph{Proceedings of the Thirtieth {AAAI} Conference on Artificial
  Intelligence, February 12-17, 2016, Phoenix, Arizona, {USA}}, pages
  3776--3784. {AAAI} Press.

\bibitem[{Shang et~al.(2018)Shang, Liu, Jiang, Ren, Voss, and Han}]{Shang_2018}
Jingbo Shang, Jialu Liu, Meng Jiang, Xiang Ren, Clare~R. Voss, and Jiawei Han.
  2018.
\newblock \href {https://doi.org/10.1109/tkde.2018.2812203} {Automated phrase
  mining from massive text corpora}.
\newblock \emph{IEEE Transactions on Knowledge and Data Engineering},
  30(10):1825–1837.

\bibitem[{Sun et~al.(2019)Sun, Tang, Du, Deng, and
  Nie}]{sun2019divgraphpointer}
Zhiqing Sun, Jian Tang, Pan Du, Zhi-Hong Deng, and Jian-Yun Nie. 2019.
\newblock Divgraphpointer: A graph pointer network for extracting diverse
  keyphrases.
\newblock \emph{arXiv preprint arXiv:1905.07689}.

\bibitem[{Sutskever et~al.(2014)Sutskever, Vinyals, and
  Le}]{sutskever2014seq2seq}
Ilya Sutskever, Oriol Vinyals, and Quoc~V. Le. 2014.
\newblock Sequence to sequence learning with neural networks.
\newblock In \emph{NIPS}.

\bibitem[{Swaminathan et~al.(2020)Swaminathan, Zhang, Mahata, Gosangi, Shah,
  and Stent}]{DBLP:conf/emnlp/SwaminathanZMGS20}
Avinash Swaminathan, Haimin Zhang, Debanjan Mahata, Rakesh Gosangi, Rajiv~Ratn
  Shah, and Amanda Stent. 2020.
\newblock \href {https://doi.org/10.18653/v1/2020.emnlp-main.645} {A
  preliminary exploration of gans for keyphrase generation}.
\newblock In \emph{Proceedings of the 2020 Conference on Empirical Methods in
  Natural Language Processing, {EMNLP} 2020, Online, November 16-20, 2020},
  pages 8021--8030. Association for Computational Linguistics.

\bibitem[{Tomokiyo and Hurst(2003)}]{tomokiyo2003language}
Takashi Tomokiyo and Matthew Hurst. 2003.
\newblock A language model approach to keyphrase extraction.
\newblock In \emph{Proceedings of the ACL 2003 workshop on Multiword
  expressions: analysis, acquisition and treatment}, pages 33--40.

\bibitem[{Wan and Xiao(2008)}]{DBLP:conf/aaai/WanX08}
Xiaojun Wan and Jianguo Xiao. 2008.
\newblock \href {http://www.aaai.org/Library/AAAI/2008/aaai08-136.php} {Single
  document keyphrase extraction using neighborhood knowledge}.
\newblock In \emph{Proceedings of the Twenty-Third {AAAI} Conference on
  Artificial Intelligence, {AAAI} 2008, Chicago, Illinois, USA, July 13-17,
  2008}, pages 855--860. {AAAI} Press.

\bibitem[{Wang et~al.(2016)Wang, Zhao, and Huang}]{wang2016ptr}
Minmei Wang, Bo~Zhao, and Yihua Huang. 2016.
\newblock Ptr: Phrase-based topical ranking for automatic keyphrase extraction
  in scientific publications.
\newblock \emph{ICONIP 2016}.

\bibitem[{Xiong et~al.(2019)Xiong, Hu, Xiong, Campos, and
  Overwijk}]{xiong2019open}
Lee Xiong, Chuan Hu, Chenyan Xiong, Daniel Campos, and Arnold Overwijk. 2019.
\newblock Open domain web keyphrase extraction beyond language modeling.
\newblock \emph{arXiv preprint arXiv:1911.02671}.

\bibitem[{Ye and Wang(2018)}]{ye_18}
Hai Ye and Lu~Wang. 2018.
\newblock Semi-supervised learning for neural keyphrase generation.
\newblock In \emph{{EMNLP}}, pages 4142--4153. Association for Computational
  Linguistics.

\bibitem[{Yuan et~al.(2018)Yuan, Wang, Meng, Thaker, Brusilovsky, He, and
  Trischler}]{yuan2018size}
Xingdi Yuan, Tong Wang, Rui Meng, Khushboo Thaker, Peter Brusilovsky, Daqing
  He, and Adam Trischler. 2018.
\newblock \href {http://arxiv.org/abs/1810.05241} {One size does not fit all:
  Generating and evaluating variable number of keyphrases}.

\bibitem[{{Zhang} et~al.(2017){Zhang}, {Fang}, and {Weidong}}]{8248519}
Y.~{Zhang}, Y.~{Fang}, and X.~{Weidong}. 2017.
\newblock Deep keyphrase generation with a convolutional sequence to sequence
  model.
\newblock In \emph{2017 4th International Conference on Systems and Informatics
  (ICSAI)}, pages 1477--1485.

\bibitem[{Zhang et~al.(2004)Zhang, Zincir-Heywood, and
  Milios}]{10.5555/1039791.1039794}
Yongzheng Zhang, Nur Zincir-Heywood, and Evangelos Milios. 2004.
\newblock World wide web site summarization.
\newblock \emph{Web Intelli. and Agent Sys.}, 2(1):39–53.

\end{thebibliography}

\end{document}